\documentclass[sigconf]{acmart}
\usepackage{booktabs}
\usepackage{multirow}
\usepackage{adjustbox}
\usepackage{graphicx}
 \usepackage{colortbl} 
\usepackage{listings}
\colorlet{punct}{black!60!black}
\definecolor{background}{HTML}{EEEEEE}
\definecolor{delim}{RGB}{20,105,176}
\colorlet{numb}{black!60!black}

\lstdefinelanguage{json}{
    basicstyle=\normalfont\ttfamily,
    numbers=left,
    numberstyle=\scriptsize,
    stepnumber=1,
    numbersep=8pt,
    showstringspaces=false,
    breaklines=true,
    frame=lines,
    extendedchars=true,
    backgroundcolor=\color{background},
    literate=
     *{0}{{{\color{numb}0}}}{1}
      {1}{{{\color{numb}1}}}{1}
      {2}{{{\color{numb}2}}}{1}
      {3}{{{\color{numb}3}}}{1}
      {4}{{{\color{numb}4}}}{1}
      {5}{{{\color{numb}5}}}{1}
      {6}{{{\color{numb}6}}}{1}
      {7}{{{\color{numb}7}}}{1}
      {8}{{{\color{numb}8}}}{1}
      {9}{{{\color{numb}9}}}{1}
      {:}{{{\color{punct}{:}}}}{1}
      {,}{{{\color{punct}{,}}}}{1}
      {\{}{{{\color{delim}{\{}}}}{1}
      {\}}{{{\color{delim}{\}}}}}{1}
      {[}{{{\color{delim}{[}}}}{1}
      {]}{{{\color{delim}{]}}}}{1}
      {á}{{\'a}}1 
      {ñ}{{\~n}}1 
      {é}{{\'e}}1
      {ó}{{\'o}}1
      {ü}{{\"u}}1 
      {ö}{{\"o}}1
      {ä}{{\"a}}1
      {ß}{{\ss}}1
      {í}{{\'i}}1,
}

\AtBeginDocument{%
  }

\begin{document}

\title{Enhancing E-commerce Product Title Translation with Retrieval-Augmented Generation and Large Language Models}


\author{Bryan Zhang}
\affiliation{%
  \institution{Amazon}
  \city{Pittsburgh}
  \country{USA}}
\email{bryzhang@amazon.com}

\author{Taichi Nakatani}
\affiliation{%
  \institution{Amazon}
  \city{Pittsburgh}
  \country{USA}}
\email{taichina@amazon.com}

\author{Stephan Walter}
\affiliation{%
 \institution{Amazon}
 \city{Berlin}
 \country{Germany}}
\email{sstwa@amazon.de}







\begin{abstract}
E-commerce stores enable multilingual product discovery which require accurate product title translation. Multilingual large language models (LLMs) have shown promising capacity to perform machine translation tasks, and it can also enhance and translate product titles cross-lingually in one step. However, product title translation often requires more than just language conversion because titles are short, lack context, and contain specialized terminology. 
This study proposes a retrieval-augmented generation (RAG) approach that leverages existing bilingual product information in e-commerce by retrieving similar bilingual examples and incorporating them as few-shot prompts to enhance LLM-based product title translation. Experiment results show that our proposed RAG approach improve product title translation quality with chrF score gains of up to 15.3\% for language pairs where the LLM has limited proficiency.
\end{abstract}



\maketitle
\section{Introduction}
As e-commerce shopping websites become localized worldwide, more customers are provided with options to browse products in their preferred language other than the primary language of the store. 
To accomplish this, modern e-commerce stores enable multi-lingual product discovery \citep{10.1145/3308558.3313502,nie2010cross,saleh-pecina-2020-document,bi2020constraint,jiang-etal-2020-cross,gartner2021} as well as localizing product information such as titles using machine translation (MT) systems \citep{way2013traditional,guha2014machine, DBLP:journals/corr/abs-1808-08266,wang2021progress,zhang-etal-2023-enhancing-arabic}. Product titles play a crucial role in conveying key details about the products, and it is essential that this information is accurately presented in the localized language. 

E-commerce product title localization traditionally uses a fleet of  bilingual neural machine translation (NMT) systems. However, recent advancements in multilingual large language models (LLMs) have demonstrated promising performance in machine translation as a single model for high resource languages \cite{hendy2023goodgptmodelsmachine}, which makes LLMs as a viable alternative to conventional NMT for translating product titles. Furthermore, LLMs have shown capabilities to optimize the length and the enrich the content of product titles within a single language, which suggest that they may be able to tackle title enhancement and translation in a cross-lingual manner as a one-step unified process \cite{zhang-etal-2024-dont}. As a result, LLMs are increasingly becoming a more prominent and integral component of e-commerce stores' product title localization strategy.

However, there are challenges in using LLMs for product title localization in e-commerce: (1) product titles tend to be short, and proper title translation also often require bilingual product-specific terminology and catalog domain knowledge.  For example, in the title \textit{``Dance your cares away - greeting card''}, the core phrase \textit{``Dance your cares away''} needs to be preserved in the translation, and the formality and style of the titles also must be maintained; (2) e-commerce is dynamic in nature, with new products rapidly emerging worldwide and requiring the model to have up-to-date product-specific knowledge; (3) a large number of language pairs need to be supported in the e-commerce domain but LLMs may not possess sufficient language proficiency for certain language pairs \cite{guerreiro2023hallucinationslargemultilingualtranslation}, which limits their effectiveness;  (4) newer and more capable LLMs are continuously arising, prompting the need to flexibly and quickly replace the LLMs used for product title localization in order to maintain high translation quality.

To address these challenges and improve product title localization in e-commerce, this study proposes a retrieval-augmented generation (RAG) approach. Our approach utilizes the constantly-growing bilingual catalog of products as source of product domain-specific knowledge and multilingual support, which is independent and provides flexibility to work with different LLMs. This method retrieves bilingual product information (e.g. titles, bulletpoints and descriptions) that are similar to the source title and incorporates them as few-shot examples in the prompt to enhance product title translation with LLMs. Finally, the study analyzes the product title translations generated from LLMs with the RAG approach to understand its impact on translation quality.

Our experiment shows that our proposed LLM-based retrieval-augmented generation (RAG) approach for product title translation can significantly enhance the title translation quality with up to 15.3\%  chrF score improvement for language pairs where the LLM does not primarily support.
\
\section{RAG approach for product title translation}

\subsection{Utilizing bilingual product information for RAG}
\begin{figure}[t]
\centering
  \includegraphics[width=0.8\linewidth]{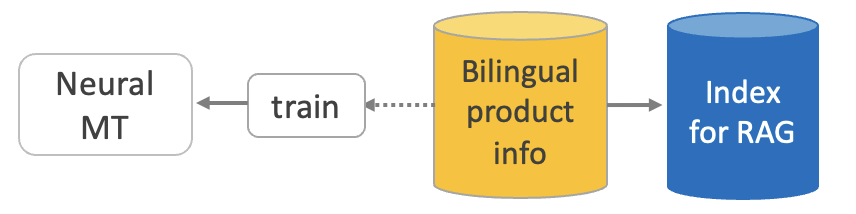}
  \caption{Index large volume of bilingual product information for RAG }  
  \label{fig:experiments}
\end{figure}

In the e-commerce industry, we can leverage the large and constantly growing volumes of bilingual product information data, including titles, bullet points, and descriptions that can be acquired.  These multilingual product titles, descriptions and bulletpoints are commonly used to train smaller neural machine translation (NMT) models for product title translation tasks. Given the dynamic nature of e-commerce, where product information localization is an ongoing process, such bilingual product content continues to expand over time. The bilingual product content can be further utilized to build a search index serving as a source of product domain-specific knowledge and multilingual support for LLM RAG approaches as illustrated in Figure \ref{fig:experiments}.
\subsection{Using retrieved similar product information as few-shot examples}
\label{sec:cptg}

\begin{figure}[t]
\centering
\includegraphics[width=1\linewidth]{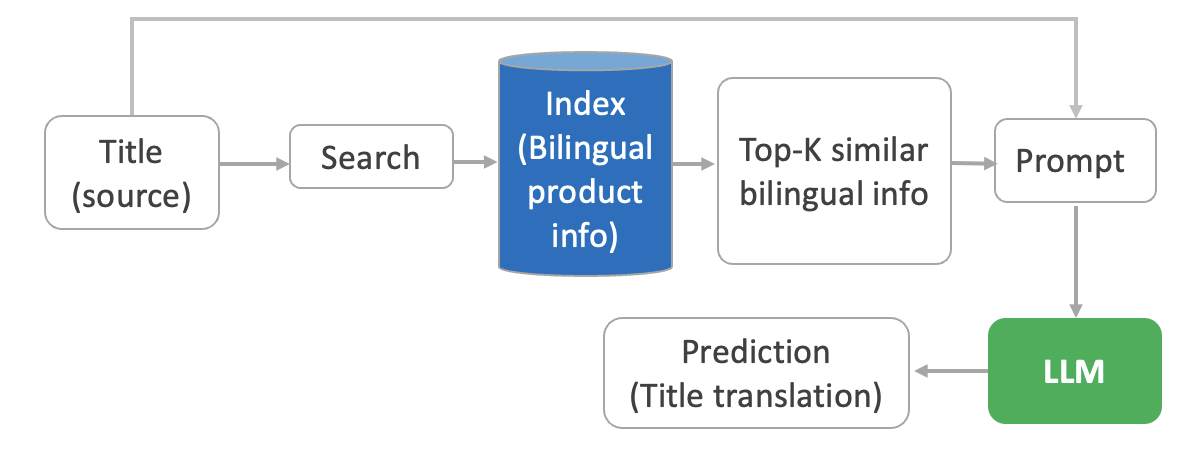}
\caption{Proposed RAG approach for product title translation using LLM}

\label{fig:experiments-2}
\end{figure}

We propose harnessing this ever-increasing bilingual product information data to power a sustainable and scalable retrieval-augmented generation (RAG) approach to enhance product title translation. By retrieving similar bilingual product information examples from this data and using them as few-shot prompts, we can guide the LLMs to generate higher quality and more contextually appropriate translations for product titles as shown in Figure \ref{fig:experiments-2}. This RAG-based approach allows us to continuously leverage the benefits of the existing and expanding bilingual product information to improve the performance of our cross-lingual title translation capabilities.

Our proposed RAG approach for product title translation is as follows.
Given a large volume of bilingual product information data (e.g. titles), $$D_{bil} = \{(t_{src}^0, t_{tgt}^0),(t_{src}^1, t_{tgt}^1) ... (t_{src}^n, t_{tgt}^n) \}$$ where $t_{src}$ and $t_{tgt}$ are product information text (e.g. titles) in the source and target languages from a bilingual language pair $L$, we propose to build an index $I_L$ on the product information (e.g. titles) in the source language and cache the corresponding target language counterparts. We recommend using text-based retrieval frameworks such as BM25, as these can better focus on textual patterns and are easy to scale and maintain, previous study has also shown that BM25 is an effective retrieval mechanism for RAG as it can achieve the high RAG performance with a number of LLMs \cite{guinet2024automatedevaluationretrievalaugmentedlanguage}. As newer products and language pairs emerge, they can be readily incorporated into the index.



At inference time, given a product title $t'$ in the source language for the language pair $L$, we retrieve from the index $I_L$ the top-$k$ bilingual product information (e.g. titles) $$D_{topk} = \{(t_{src}^0, t_{tgt}^0),(t_{src}^1, t_{tgt}^1) ... (t_{src}^k, t_{tgt}^k) \}$$  based on the similarity scoring $S(t', t_{src})$ of the source product information (e.g. titles) where,  $D_{topk}\subset D_{bil}$,  $S(t', t_{src}^0) \geq S(t', t_{src}^1)\geq ...\geq S(t', t_{src}^k)\geq ...\geq S(t', t_{src}^n)$ and $n=|I_L|=|D_{bil}|$. We then integrate these retrieved top-$k$ bilingual product information (e.g. titles) into the prompts as few-shot examples. The retrieved product information (e.g. titles) share similar textual patterns, language, and semantics with the given title, allowing the LLM to leverage these references to generate a translation that is more aligned with the desired objectives.

The following is the prompt template A for title translation task without RAG, brackets (\texttt{<>}) are placeholders and substituted with relevant text.
\begin{lstlisting}[language=json,firstnumber=1,basicstyle=\scriptsize,numbers=none]
Below is a source text, it is a title of a product in <source language e.g. English>: <title in the source language> 
Your task is to translate the source text into <target language>. Here are instructions on how to translate: 
- Overall, make sure the translation is accurate, particularly pay attention to the terminology. 
- Brands are generally preserved in the translation. If you are very confident there is a different brand term in <target language>, use it in the translation. 

Return your response in JSON format with the 'translation' key containing your translation as {"translation": <translation>} Please make sure the output is in JSON format, don't add other information.
\end{lstlisting}

The following is the prompt template B for title translation task with few-shot examples, we integrate the top-$k$ retrieved similar bilingual title as few-shot examples in the prompt. The brackets (\texttt{<>}) are placeholders and substituted with relevant text.
\begin{lstlisting}[language=json,firstnumber=1,basicstyle=\scriptsize,numbers=none]
Below is a source text, it is a title of a product in <source language e.g. English>: 
<title in the source language> 
Your task is to translate the source text into <target language>.
Below are a few translaton examples:
    Example 1: source: <source title>, 
               translation: <title translation>
    Example 2: source: <source title>, 
               translation: <title translation>
    Example 3: source: <source title>, 
               translation: <title translation>
    Example 4: source: <source title>, 
               translation: <title translation>
    Example 5: source: <source title>, 
               translation: <title translation>

Here are instructions on how to translate: 
- Overall, make sure the translation is accurate, particularly pay attention to the terminology. 
- Brands are generally preserved in the translation. If you are very confident there is a different brand term in <target language>, use it in the translation. 
- Refer to the provided translation examples if you could find any useful information. For example:  any terminology translations, phrases translation, how the brand term is handled etc.  only if you find it useful!

Return your response in JSON format with the 'translation' key containing your translation as {"translation": <translation>} 
Please make sure the output is in JSON format, don't add other information.
\end{lstlisting}

\section{Experiment}
\noindent\textbf{Language pairs}: We experiment with 7 language pairs  English-Dutch (EN-NL), English-German (EN-DE), Italian-German (IT-DE), Turkish-German (TR-DE), German-Czech (DE-CS), English-Polish (EN-PL) and English-Swedish (EN-SE). \\

\noindent\textbf{Test data and MT metric}: For each language pair, we sample 2000 product titles as our test set and evaluate translation quality using chrF \citep{popovic-2015-chrF}.\footnote{We use chrF instead of BLEU because the average length of a title in our testsets is 11.3 raw words, making BLEU's four-gram statistics sparse and unreliable even at the corpus level \cite{kocmi-etal-2021-ship} while the chrF operates at a character level which is more suitable for short texts such as titles.} \\

\noindent\textbf{RAG framework}: We use \texttt{Lucene 8.11.3} and OKAPI BM25 probabilistic retrieval framework to build each index. For each bilingual product information data pair, we index the source text and cache the target text. \\

\noindent\textbf{RAG data}: We sample large volumes of bilingual product (titles, descriptions and bulletpoints) for each language pair to build the index for our RAG approach. For each language pair, we build four indices using sampled bilingual product information data of different domains: product titles (TTL), bullet points (BP), descriptions (PD) and a combination of aforementioned domains (T.B.D.).  Language pairs EN-NL, EN-PL, EN-DE and IT-DE have over 10 million bilingual pairs in their T.B.D. index, and the rest have millions.  \\

\noindent\textbf{Experiment configurations}:
We set up the following experiment configurations using different prompts for title translation generation with LLM:\\
\noindent \textit{Baseline Generation}: The prompt using the \textit{prompt template A} presented in section \ref{sec:cptg}, which does not incorporate retrieved or randomly sampled few-shot examples.\\

\noindent \textit{RAND 1-shot} and \textit{RAND 5-shot}: The prompt using the \textit{prompt template B} presented in section \ref{sec:cptg}, which incorporates 1 and 5 randomly-chosen examples from the respective domain (TTL, BP, PD, T.B.D.) as few-shot prompts.  These serve as the lower-threshold of our RAG approach.\\

\noindent \textit{RAG 1-shot} and \textit{RAG 5-shot}: The prompt using the \textit{prompt template B} in section \ref{sec:cptg}, which incorporates 1 and 5 most similar examples retrieved from their respective domain index (TTL, BP, PD, T.B.D.) as few-shot prompts.\\

\noindent\textbf{LLM}: We use \texttt{Mixtral-8x7B-Instruct}\footnote{\url{https://huggingface.co/mistralai/Mixtral-8x7B-Instruct-v0.1}}, a publicly available LLM, to translate source product title into the target language through prompting using the \textit{prompt template} presented in section \ref{sec:cptg}. \\

\section{Results and Analysis}
\begin{table}[ht]
\centering
\begin{adjustbox}{width=0.45\textwidth}

\begin{tabular}{@{}clrrrr@{}}
\toprule
\textbf{}                  &    & \textbf{TTL }      & \textbf{BP}        & \textbf{PD}        & \textbf{T.B.D.}                \\ \midrule
\textbf{EN-PL}             &             & \textbf{}            & \textbf{}            & \textbf{}            & \textbf{}                     \\
\multirow{4}{*}{\textbf{}} & RAND 1-shot & -1.0\%               & -0.7\%               & +0.2\%               & -0.4\%                        \\
                           & RAND 5-shot & -1.7\%               & -0.8\%               & -3.0\%               & -2.6\%                        \\
                           & RAG 1 shot  & +8.3\%              & +2.6\%               & +10.7\%              & +10.9\%            \\
                           & RAG 5 shot  & +11.8\%              & +4.6\%               & +14.9\%              & \underline{\textbf{+15.3\%}}                       \\\midrule
\textbf{EN-NL}             &             &                      &                      &                      &                               \\
\multirow{4}{*}{\textbf{}} & RAND 1-shot & +1.9\%               & +1.1\%               & +3.4\%               & +2.9\%                        \\ 
                           & RAND 5-shot & +1.7\%               & +2.0\%               & +5.1\%               & +4.8\%                        \\
                           & RAG 1 shot  & +3.0\%               & +2.3\%               & +7.2\%              & +7.4\%             \\
                           & RAG 5 shot  & +6.2\%               & +3.3\%               & \underline{\textbf{+12.0\%}}     & +11.9\%                       \\\midrule
\textbf{EN-SE}             &             &                      &                      & \textbf{}            &                               \\
\multirow{4}{*}{\textbf{}} & RAND 1-shot & +1.8\%               & +1.6\%               & +0.9\%               & +1.6\%                        \\ 
                           & RAND 5-shot & +2.8\%               & +2.4\%               & +1.7\%               & +2.4\%                        \\
                           & RAG 1 shot  & +8.8\%              & +4.6\%               & +11.2\%     & +10.8\%                        \\
                           & RAG 5 shot  & +11.5\%              & +6.4\%               & \underline{\textbf{+14.0\%}}             & +6.4\%                        \\\midrule
\textbf{DE-CS}             &             & \multicolumn{1}{l}{} & \multicolumn{1}{l}{} & \multicolumn{1}{l}{} & \multicolumn{1}{l}{}          \\
\multirow{4}{*}{\textbf{}} & RAND 1-shot & -1.0\%               & -0.2\%               & -1.7\%               & -2.6\%                        \\
                           & RAND 5-shot & -2.4\%               & +1.3\%               & +0.0\%               & -1.4\%                        \\
                           & RAG 1 shot  & +3.2\%               & +2.0\%               & +1.0\%               & +3.2\%               \\
                           & RAG 5 shot  & +5.3\%               & +4.3\%               & +2.7\%               & \underline{\textbf{+5.9\%}}                       \\ \midrule
\textbf{EN-DE}             &             & \multicolumn{1}{l}{} & \multicolumn{1}{l}{} & \multicolumn{1}{l}{} & \multicolumn{1}{l}{\textbf{}} \\
\multirow{4}{*}{\textbf{}} & RAND 1-shot & -1.1\%               & -0.9\%               & -0.9\%               & -0.8\%                        \\ 
                           & RAND 5-shot & -0.6\%               & -1.7\%               & -4.0\%               & -0.7\%                        \\
                           & RAG 1 shot  & +1.5\%     & +1.1\%               & -0.9\%               & +0.9\%                        \\
                           & RAG 5 shot  &\underline{\textbf{ +2.0\%}}     & +1.4\%               & -1.6\%               & +1.7\%                        \\\midrule
\textbf{IT-DE}             &             & \multicolumn{1}{l}{} & \multicolumn{1}{l}{} & \multicolumn{1}{l}{} & \multicolumn{1}{l}{}          \\
\multirow{4}{*}{\textbf{}} & RAND 1-shot & +1.8\%               & +0.0\%               & +1.0\%               & +1.0\%                        \\
                           & RAND 5-shot & +0.2\%               & +0.0\%               & +0.5\%               & +0.3\%                        \\
                           & RAG 1 shot  & +0.1\%               & +0.2\%               & +2.6\%               & +2.7\%              \\
                            & RAG 5 shot  & -0.6\%               & +0.3\%               & +3.4\%               & \underline{\textbf{+3.6\%}}                        \\\midrule
\textbf{TR-DE}             &             &                      &                      &                      &                               \\
\multirow{4}{*}{\textbf{}} & RAND 1-shot & +0.3\%               & +1.1\%               & +0.5\%               & +0.3\%                        \\ 
                           & RAND 5-shot & +0.7\%               & +0.2\%               & +0.1\%               & +1.2\%                        \\
                           & RAG 1 shot  & +1.7\%     & +1.4\%               &+1.6\%               & \underline{\textbf{+2.3\%}}                        \\
                           & RAG 5 shot  & +1.6\%               & +0.9\%               & +0.5\%               & +1.4\%                        \\\midrule
\end{tabular}
\end{adjustbox}
\caption{$\Delta$ chrF\% of title translation experiments using against baseline generation. ``RAG \{1,5\} shot'' indicate top-1 and 5 retrieved example(s) based on Okapi BM25 similarity score. ``RAND \{1,5\} shot'' indicates 1 and 5 randomly chosen example(s).  TTL/BP/PD/T.B.D. refer to index built using bilingual titles, bulletpoints, descriptions, and all three domains respectively.} 
\label{tab:rag_result}
\end{table}

Table \ref{tab:rag_result} shows the chrF deltas relative to the \textit{baseline generation}. We observe that RAG 1-shot and RAG 5-shot overall improve title translation quality across different data index domains (TTL, BP, PD, and T.B.D.) used for retrieving few-shot examples. In contrast, configurations which retrieve random examples (RAND 1-shot and RAND 5-shot) generally show decreases in chrF, which indicate a negative impact on title translation quality.  Where improvements are seen amongst RAND 1 and 5-shot configurations, the magnitude is much smaller than that of the RAG 1-shot and RAG 5-shot configurations.

Results show RAG 5-shot with examples from all domains (T.B.D. index) has the largest or second largest title translation quality improvement across all language pairs except EN-SE.
For language pairs where the target language is not the main language of the LLM, RAG 5-shot demonstrate chrF improvements ranging from +5.9\% to +15.3\% across DE-CS, EN-SE, EN-NL, and EN-PL. For TR-DE, where the source language is not a main language of the LLM, RAG 1-shot show improvements by up to 2.3\% chrF.


RAG 1-shot and RAG 5-shot configurations also demonstrate improvements in chrF for language pairs where both languages are the main supported languages of the LLM. Those language pairs have higher chrF for baseline generation, and also have chrF improvement up to 3.6\% for IT-DE and 2.0\% for EN-DE.

Notably, all language pairs show the greatest increase in chrF when examples are retrieved from the T.B.D. index which contains all three product information domains (title, bullet points, and description). This suggests that a combined index can provide more useful contextual cues to guide the LLM's product title translation task.\\

\begin{table}[h]
\begin{adjustbox}{width=0.45\textwidth}
\begin{tabular}{@{}ll@{}}
\toprule
\multicolumn{2}{l}{\textbf{Example A (EN-DE)}} \\ \midrule
\multicolumn{1}{l|}{\textit{src}} & \begin{tabular}[c]{@{}l@{}}Posted No Trespass Hunting Fishing Trapping Under \\ Penalty 10x7 Aluminum  sign\end{tabular} \\ \midrule
\multicolumn{1}{l|}{\textit{ref}} & \begin{tabular}[c]{@{}l@{}}``Posted No Trespass Hunting Fishing Trapping Under \\ Penalty ''   25,4 x 17,8 cm Aluminum-Schild\end{tabular} \\ \midrule
\multicolumn{1}{l|}{\textit{bl}} & \begin{tabular}[c]{@{}l@{}}Angebrachtes No Trespassing-Schild für Jagd, Fischfang \\ und Fallenstellen   unter Strafe 10x7 Aluminium\end{tabular} \\ \midrule
\multicolumn{1}{l|}{\textit{rag5}} & \begin{tabular}[c]{@{}l@{}}Aluminiumschild Posted No Trespass Hunting Fishing \\ Trapping under Penalty 10x7, Metall\end{tabular} \\ \midrule
\rowcolor[HTML]{EFEFEF} 
\multicolumn{1}{l|}{\cellcolor[HTML]{EFEFEF}\textit{rag src}} & \begin{tabular}[c]{@{}l@{}}No Hunting Sign - Private Property Posted Hunter Keep\\  Out Signs -   Aluminum Metal\end{tabular} \\ \midrule
\rowcolor[HTML]{EFEFEF} 
\multicolumn{1}{l|}{\cellcolor[HTML]{EFEFEF}\textit{rag tgt}} & \begin{tabular}[c]{@{}l@{}}Schild ``No Hunting'', für Privatbesitz, um Jäger abzuhalten,   \\ Aluminium, Metall\end{tabular} \\ \midrule
\multicolumn{2}{l}{\textbf{Example B (EN-DE)}} \\ \midrule
\multicolumn{1}{l|}{\textit{src}} & KYB Shock Absorber, Part Number: 343441 \\ \midrule
\multicolumn{1}{l|}{\textit{ref}} & KYB Stoßdämpfer, Teilenummer: 343441 \\ \midrule
\multicolumn{1}{l|}{\textit{bl}} & Dämpferstabilisator KYB,   Teilenummer: 343441 \\ \midrule
\multicolumn{1}{l|}{\textit{rag5}} & KYB Stoßdämpfer, Teilenummer:   343441 \\ \midrule
\rowcolor[HTML]{EFEFEF} 
\multicolumn{1}{l|}{\cellcolor[HTML]{EFEFEF}\textit{rag src}} & KYB 554105, KYB Front Shock Absorber (Single Unit) 554105 \\ \midrule
\rowcolor[HTML]{EFEFEF} 
\multicolumn{1}{l|}{\cellcolor[HTML]{EFEFEF}\textit{rag tgt}} & KYB 554105, KYB-Stoßdämpfer vorne   (Einzeleinheit) 554105 \\ \midrule
\multicolumn{2}{l}{\textbf{Example C (EN-DE)}} \\ \midrule
\multicolumn{1}{l|}{\textit{src}} & \begin{tabular}[c]{@{}l@{}}mumbi ultra slim case for Nokia 6 2018, protective case \\ in mandala design, transparent pink (ultra slim 0.70 mm)\end{tabular} \\ \midrule
\multicolumn{1}{l|}{\textit{ref}} & \begin{tabular}[c]{@{}l@{}}mumbi UltraSlim Hülle für Nokia 6 2018 Schutzhülle im\\  Mandala-Design,   transparent rosa (Ultra Slim - 0,70 mm)\end{tabular} \\ \midrule
\multicolumn{1}{l|}{\textit{bl}} & \begin{tabular}[c]{@{}l@{}}Hülle für Nokia 6 2018 von mumbi, ultra schlanker Schutzcase mit  \\ Mandala-Muster, transparente Rosatönung (ultra dünn 0,70 mm)\end{tabular} \\ \midrule
\multicolumn{1}{l|}{\textit{rag5}} & \begin{tabular}[c]{@{}l@{}}mumbi Ultra Slim Hülle für Nokia 6 von 2018, Schutzhülle\\ im Mandala   Design, transparente Rosa (ultra slim 0,70 mm)\end{tabular} \\ \midrule
\rowcolor[HTML]{EFEFEF} 
\multicolumn{1}{l|}{\cellcolor[HTML]{EFEFEF}\textit{rag src}} & \begin{tabular}[c]{@{}l@{}}mumbi Ultra-Slim Phone Case Ultra-Slim for Huawei P9 \\ Lite Case in Mandala   Design Transparent Pink 0.55 mm.\end{tabular} \\ \midrule
\rowcolor[HTML]{EFEFEF} 
\multicolumn{1}{l|}{\cellcolor[HTML]{EFEFEF}\textit{rag tgt}} & \begin{tabular}[c]{@{}l@{}}mumbi UltraSlim Hülle für Huawei P9 Lite Schutzhülle im \\ Mandala Design   transparent rosa (Ultra Slim - 0.55 mm)\end{tabular} \\ \midrule
\end{tabular}
\end{adjustbox}
\centering
\caption{Translation examples using LLM with RAG (\textit{rag5}) and baseline generation (\textit{bl}) with no RAG. \textit{rag src} and \textit{rag tgt} refer to the retrieved top-one example used as few-shot in the prompt. \textit{src}, \textit{ref} refer to the title in the source language and reference title translation respectively  }
\label{tab:rag-example}
\end{table}
\begin{table}[h]
\centering
\begin{adjustbox}{width=0.4\textwidth}
\begin{tabular}{@{}llrrrr@{}}
\toprule
 & \multicolumn{1}{c}{} & \multicolumn{1}{r}{\textbf{TTL}} & \multicolumn{1}{r}{\textbf{BP}} & {\textbf{PD}} & {\textbf{T.B.D}} \\ \midrule
 \textbf{EN-PL} & \textbf{} &  &  &  &  \\
  & RAND 1-shot & 14.3 & 11.0 & \underline{\textbf{16.4}} & 11.0 \\
 & RAND 5-shot & 13.0 & 11.6 & \underline{\textbf{15.0}} & 13.2 \\
 & RAG 1-shot & 46.4 & 35.2 & \underline{\textbf{52.4}} & 49.1 \\
 & RAG 5-shot & 40.9 & 32.6 & \underline{\textbf{43.6}} & 39.0 \\\midrule
 \textbf{EN-NL} &  &  &  &  &  \\
  & RAND 1-shot & \underline{\textbf{16.2}} & 11.1 & 4.0 & 4.0 \\
 & RAND 5-shot & \underline{\textbf{15.7}} & 14.2 & 10.2 & 13.3 \\
 & RAG 1-shot & 42.5 & 32.0 & \underline{\textbf{54.2}} & 42.6 \\
 & RAG 5-shot & 38.0 & 29.2 & \underline{\textbf{45.2}} & 37.5 \\\midrule
\textbf{EN-SE} & \textbf{} & \multicolumn{1}{l}{} & \multicolumn{1}{l}{} & \multicolumn{1}{l}{} & \multicolumn{1}{l}{} \\
  & RAND 1-shot & 6.1 & 6.2 & \underline{\textbf{6.3}} & 6.1 \\
 & RAND 5-shot & 8.4 & \underline{\textbf{8.5}} & 8.3 & 8.4 \\
 & RAG 1-shot & 46.4 & 35.6 & \underline{\textbf{52.6}} & 47.8 \\
 & RAG 5-shot & 40.9 & 32.9 & \underline{\textbf{44.1}} & 39.3 \\\midrule
\textbf{DE-CS} & \textbf{} & \multicolumn{1}{l}{} & \multicolumn{1}{l}{} & \multicolumn{1}{l}{} & \multicolumn{1}{l}{} \\
  & RAND 1-shot & 12.2 & 10.0 & \underline{\textbf{12.9}} & 10.5 \\
 & RAND 5-shot & \underline{\textbf{12.5}} & 8.1 & 10.8 & 10.6 \\ 
 & RAG 1-shot & \underline{\textbf{37.9}} & 37.1 & 35.1 & 32.1 \\
 & RAG 5-shot & 32.1 & 31.6 & \underline{\textbf{31.6}} & 29.3 \\\midrule
 \textbf{EN-DE} & \textbf{} & \multicolumn{1}{l}{} & \multicolumn{1}{l}{} & \multicolumn{1}{l}{} & \multicolumn{1}{l}{} \\
 & RAND 1-shot & \underline{\textbf{13.3}} & 10.2 & 4.4 & 10.2 \\
 & RAND 5-shot & \underline{\textbf{13.2}} & 10.4 & 7.9 & 10.5 \\ 
 & RAG 1-shot & \underline{\textbf{49.6}} & 41.2 & 24.6 & 41.1 \\
 & RAG 5-shot & \underline{\textbf{43.1}} & 38.1 & 23.4 & 34.9 \\\midrule
 \textbf{IT-DE} & \textbf{} &  &  &  &  \\
  & RAND 1-shot & \underline{\textbf{16.6}} & 16.4 & 14.2 & 15.8 \\
 & RAND 5-shot & \underline{\textbf{14.8}} & 12.5 & 13.7 & 13.7 \\
 & RAG 1-shot & 45.7 & 42.5 & \underline{\textbf{49.2}} & 46.5 \\
 & RAG 5-shot & 40.8 & 39.3 & \underline{\textbf{41.5}} & 40.5 \\ \midrule
\textbf{TR-DE} & \textbf{} &  &  &  &  \\
 & RAND 1-shot & 10.6 & 9.4 & 9.0 & \underline{\textbf{10.7}} \\
 & RAND 5-shot & \underline{\textbf{11.0}} & 8.9 & 7.4 & 9.5 \\
 & RAG 1-shot & 27.5 & 27.9 & \underline{\textbf{29.9}} & 25.3 \\
 & RAG 5-shot & 23.9 & 23.8 & \underline{\textbf{26.9}} & 23.8 \\\bottomrule

\end{tabular}
\end{adjustbox}
\caption{Textual similarity (chrF) between the source titles and the source few-shot examples,  TTL/BP/PD/T.B.D. refer to index built using bilingual titles, bulletpoints, descriptions, and all three domains respectively.}
\label{tab:sim-rag}
\end{table}

\noindent\textbf{Retrieved examples similarity}: We compute the textual similarity between the test source title and examples used as few-shot examples in the prompts as shown in Table \ref{tab:sim-rag}. Textual similarity is computed using chrF. For RAND-5 and RAG-5 configurations, we report the average of chrF scores across the 5 examples.  We observe (1) RAG-1 and RAG-5 retrieved examples with higher similarity scores compared to RAND-1 and RAND-5 which use randomly chosen examples; (2) Given a language pair and one configuration, we observe generally that the RAG data domain with higher similarity score has larger translation quality improvement.  However, we also observe RAG-5 with examples from domains T.B.D. do not have highest similarity scores but have the largest or second largest title translation quality improvement for all language pairs except EN-SE.  This indicates that retrieved examples from more diverse product information domains can also have positive impact on the title translation generation improvement.  \\

\noindent\textbf{Title translation improvement}: Table \ref{tab:rag-example} shows examples of generated titles by the RAG LLM method.  The three examples show how the approach can improve title translation quality: \\


\noindent\textbf{(a) Improved title translation from better understanding of products}: Many product titles for products such as signs, T-shirts, mugs have text printed on the products, and such texts need to stay intact in the translation. 

Example A in Table \ref{tab:rag-example} is for a sign with the message \textit{``Posted No Trespass Hunting Fishing Trapping Under Penalty''}.  
In this example, the message should not be translated since the product includes the phrase verbatim.  Without RAG, the LLM translates the entire title to German including the phrase.  With RAG, the retrieved example is for a similar product which illustrates that the message \textit{``No Hunting''} should be preserved in the translation, and the generated title preserves the message in the German translation.  

A similar issue also occurs with named entities such as character names, brands etc., where some named entities may overlap with the common vocabulary.  An example of such case is character name \textit{``Peppa Pig''} in the product title \textit{``Peppa Pig House Tea Playset''}. Without RAG, the LLM translates it into \textit{``Peppa Schwein''} (German) while the character name is preserved in the translation with RAG. \\

\noindent\textbf{(b) Improved translation for product-specific terminology}: For example B in Table \ref{tab:rag-example}, the expected German translation of the term \textit{``Shock Absorber''} is \textit{``Stoßdämpfer''}. Without the RAG approach, the LLM instead chooses a less appropriate translation \textit{``Dämpferstabilisator''}. However, when the top-k retrieved examples contain the mapping of \textit{``Shock Absorber''} to \textit{``Stoßdämpfer''}, the LLM with RAG can then select the expected and more accurate term in the translation.\\

\noindent\textbf{(c) Improved formality and style of title translation}: We have also observed that the RAG approach can improve title quality by conforming to product title formality and style preferred in catalogs. In example C in Table \ref{tab:rag-example}, the retrieved example is for a phone case of the same brand but for a different model. The generated title translation conforms to style conventions used in the retrieved example, such as (1) placing the brand name at the beginning of the title, (2) appending the specific phone model after the brand name, and (3) including details like color. By leveraging these stylistic cues from the retrieved bilingual title pairs, the LLM can generate translations that not only convey the accurate meaning but also adhere to the preferred formatting and conventions expected in product catalogs. This helps ensure a consistent and professional-looking title presentation across the multilingual offerings.\\


\section{Related work}
Retrieval Augmented Generation (RAG) has proved to be an effective prompting strategy to enhance the accuracy and reliability of large language models (LLMs) \cite{gao2024retrievalaugmentedgenerationlargelanguage,jiang2024piperagfastretrievalaugmentedgeneration,su2024dragindynamicretrievalaugmented}. This is due to the inherent limitations of LLMs in accessing specialized knowledge \cite{ram2023incontextretrievalaugmentedlanguagemodels}. RAG enables LLMs to combine their generative abilities with additional context and factual information retrieved from external knowledge sources, which has been explored in several prior studies \cite{guu2020realmretrievalaugmentedlanguagemodel,agrawal2022incontextexamplesselectionmachine,jiang2023activeretrievalaugmentedgeneration,luo2023sailsearchaugmentedinstructionlearning,shi2023replugretrievalaugmentedblackboxlanguage}.

Retrieval-augmented methods have been successfully applied to various tasks using LLMs with domain-specific specialized knowledge, such as question answering \cite{guinet2024automatedevaluationretrievalaugmentedlanguage,hsia2024raggedinformeddesignretrieval,10.1162/tacl_a_00530}. Previous work has also proposed using RAG with a ``text book'' of retrieved language-specific usage, syntax, and vocabulary examples to improve translation tasks with LLMs \cite{guo-etal-2024-teaching-large}. However, their study focused on lower-resource language pairs, and to the best of our knowledge, there has been no prior work on applying RAG techniques to machine translation of e-commerce product titles and languages commonly used in the e-commerce industry.

\section{Conclusion}
This study presents a retrieval-augmented generation (RAG) approach to enhance e-commerce product title translation using large language models (LLMs). By leveraging the growing bilingual catalog of product content, RAG retrieves similar examples as few-shot prompts to guide the LLM in translating product terminology more accurately and generating higher quality, contextually appropriate translations. Experiments across 7 language pairs demonstrated the RAG approach's effectiveness, achieving chrF score improvements up to 15.3\%, especially for language pairs where the LLM has limited proficiency. The RAG approach can enhance translation by more accurately translating or preserving product-specific terminology, maintaining brands, and adhering to expected formatting and deliver product tile translations essential for multilingual product discovery in e-commerce. This scalable, sustainable approach also allows rapid adaptation as e-commerce continuously introduces new products and language pairs.

\bibliographystyle{ACM-Reference-Format}
\bibliography{custom-1,acl2021,anthology,custom, RAG}

\appendix

\end{document}